\documentclass[a4paper]{jpconf}
\usepackage{graphicx}
\usepackage[hidelinks]{hyperref}

\begin{document}
\title{Thermal transmittance prediction based on the application of artificial neural networks on heat flux method results}

\author{S. Gumbarević, B. Milovanović, M. Gaši and M. Bagarić}

\address{University of Zagreb, Faculty of Civil Engineering, Department of Materials, Fra Andrije Kačića Miošića 26, 10000 Zagreb, Croatia}

\ead{sanjin.gumbarevic@grad.unizg.hr, bojan.milovanovic@grad.unizg.hr, mergim.gasi@grad.unizg.hr, marina.bagaric@grad.unizg.hr}

\begin{abstract}
Deep energy renovation of building stock came more into focus in the European Union due to energy efficiency related directives. Many buildings that must undergo deep energy renovation are old and may lack design/renovation documentation, or possible degradation of materials might have occurred in building elements over time. Thermal transmittance (i.e. U-value) is one of the most important parameters for determining the transmission heat losses through building envelope elements. It depends on the thickness and thermal properties of all the materials that form a building element. In-situ U-value can be determined by ISO 9869-1 standard (Heat Flux Method – HFM). Still, measurement duration is one of the reasons why HFM is not widely used in field testing before the renovation design process commences. This paper analyzes the possibility of reducing the measurement time by conducting parallel measurements with one heat-flux sensor. This parallelization could be achieved by applying a specific class of the Artificial Neural Network (ANN) on HFM results to predict unknown heat flux based on collected interior and exterior air temperatures. After the satisfying prediction is achieved, HFM sensor can be relocated to another measuring location. Paper shows a comparison of four ANN cases applied to HFM results for a measurement held on one multi-layer wall – multilayer perceptron with three neurons in one hidden layer, long short-term memory with 100 units, gated recurrent unit with 100 units and combination of 50 long short-term memory units and 50 gated recurrent units. The analysis gave promising results in term of predicting the heat flux rate based on the two input temperatures. Additional analysis on another wall showed possible limitations of the method that serves as a direction for further research on this topic.
\end{abstract}

\section{Introduction}
The building sector is recognized as an area where action can be taken through energy efficiency regulations to reduce humanity's impact on the environment \cite{Laaroussi2020}, it is clear that there is a significant need for energy retrofitting of the building stock \cite{Jensen2018}. One of the most important criteria to increase the energy efficiency of a building is to limit the thermal transmittance (U-value) of a particular building element. It is a great challenge to estimate the U-value of the envelope element of an existing building \cite{BienvenidoHuertas2019,Ficco2015} due to the possible deterioration of the thermal properties of the layers forming the element, the influence of heterogeneity in the element, the influence of moisture in the element \cite{Gomes2017,Lucchi2017} and information loss for element materials and its properties. Therefore, in case of such assessment, it is more reasonable to use some experimental method to determine element’s U-value in existing buildings. One of common used is heat flux method (HFM) and it is standardized according to ISO 9869-1 \cite{ISO9869}, which uses a sensor to measure heat flux and two thermocouples to measure indoor and outdoor air temperatures. 
Based on the measured two temperatures and one heat flux, the U-value is estimated using one of the two proposed standardized methods -- from the average of the measured quantities (steady-state method) or using transformations of the heat conduction equation (dynamic method) \cite{BienvenidoHuertas2019,Soares2019}. In order for the results of the field test to be relevant, it is necessary to perform the test when the element is located at the boundary between the internal and external environment under conditions with a high temperature gradient \cite{BienvenidoHuertas2020}. If these conditions are not present, then there is a high probability of obtaining a significant deviation with respect to the designed U-value \cite{Tejedor2020}. When these conditions are met, the maximum deviation of the HFM result is usually limited to a maximum value of about 20\% of the difference between the designed and field U-values \cite{Evangelisti2019,Rezvani2019}. The reliability of HFM results is also affected by climate conditions of the external environment (rain, sun) and large temperature variations \cite{Meng2017}. It is especially important to keep the internal boundary condition as close as possible to a constant value \cite{Roque2020}. Moreover, the main disadvantage of HFM is the duration of the test, which can be extended to 2 weeks if the criteria of the standard are met \cite{BienvenidoHuertas2019,Meng2017,Roque2020}. The reliability of the estimated U-value is significantly increased when the dynamic method is used in the post-processing of the results \cite{Gasi2019,Choi2017}, since for the six analyzed cases in \cite{Choi2017} there is a difference of about 10\% compared to the designed U-value when using the steady-state method, and this difference when using the dynamic method does not exceed 3\%. When using the dynamic method, it is possible to reduce the testing time by one or two days if we consider the three analyzed cases in \cite{Gaspar2018}, but the measurement can still take up to 7 days. Since the boundary conditions are difficult to handle, as the external boundary condition is the outside air temperature and the internal one is the constant room temperature \cite{Roque2020}, the test time of several different building elements can be reduced either by using multiple HFM sensors (which is financially and operationally inefficient \cite{Marquez2017}) or by evaluating the sensor results according to the assumed behavior pattern. Such prediction of sensor results is potentially possible with artificial intelligence \cite{Gumbarevic2020}.

\section{Methodology}
Methods and instruments used in this research can be classified in two groups -- methods and instruments for data collection and methods for data prediction. For data collection, HFM is used for collection of internal and external air temperatures as well as heat flux through specialized sensor. HFM is deeply explained in section \ref{s:hfm}. For data analysis and prediction, artificial neural network methods are used and they are explained in section \ref{s:ann}. 

\subsection{Heat flux method}\label{s:hfm}

HFM is (as stated before) standardized method for measuring heat flux through the building element and for estimating the field U-value based on measured parameters. HFM instrument kit that is used in the research is gSKIN{\textregistered}Heat Flux Sensor.

The goal of this research is to calculate the U-value using several ANN models and to find one that has the best performance in a compare to the U-value calculated using the average method defined in the ISO 9869-1 standard \cite{ISO9869} (\ref{eq:1}):

\begin{equation} \label{eq:1}
	U = \frac{\sum_{j=1}^{N} \, q_j}{\sum_{j=1}^{N} \, (T_i - T_e)_j} \, ,
\end{equation}

\noindent where \emph{T$_i$} and \emph{T$_e$} are the internal and external air temperatures, respectively, and q is the heat flow rate. Index $j$ is corresponding to the measurement data point.

\begin{figure}[h]
	\includegraphics[width=16pc]{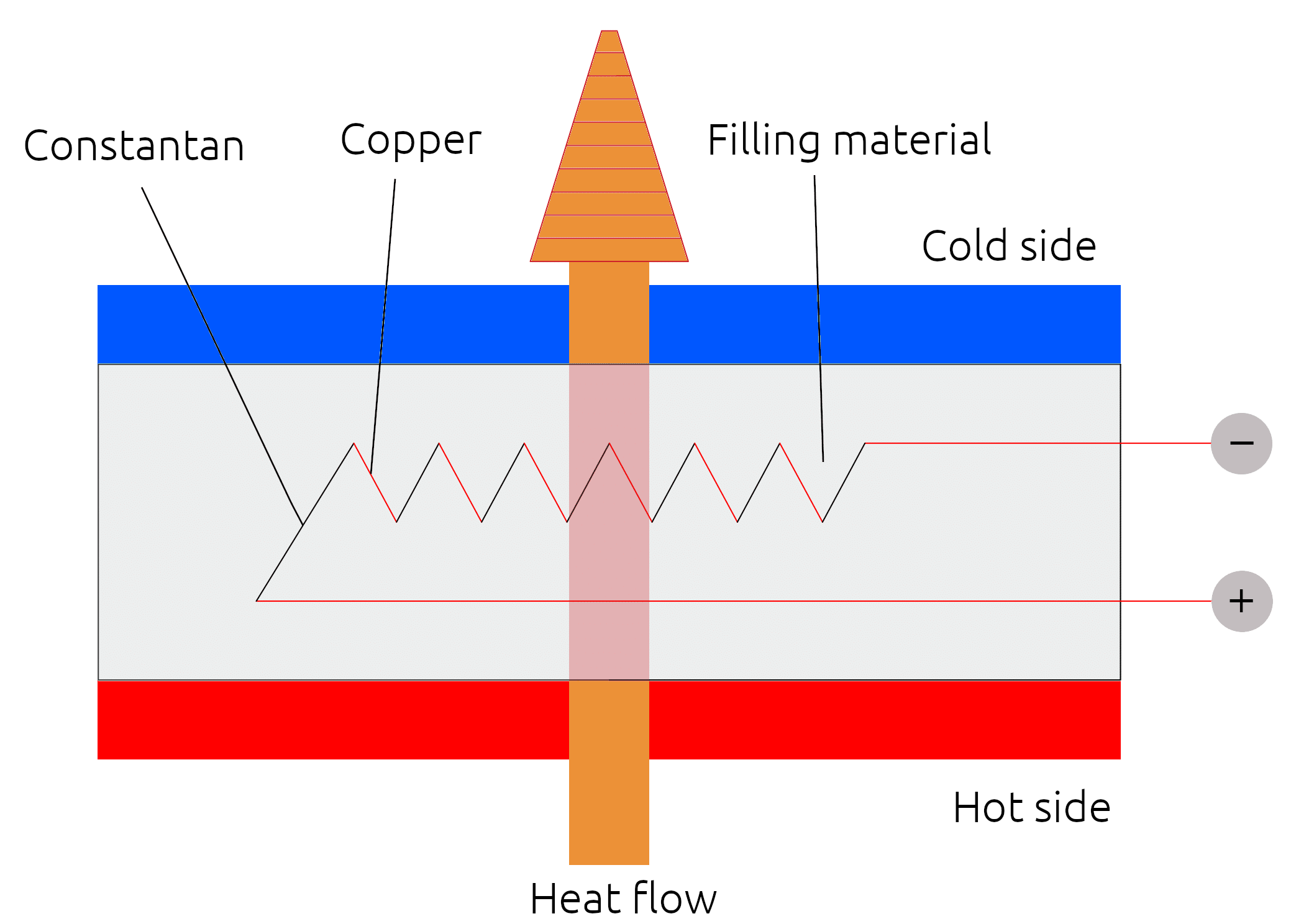}\hspace{2pc}%
	\begin{minipage}[b]{20pc}\caption{\label{f:hfm}Schematic cross section through heat flux meter.}
	\end{minipage}
\end{figure}

The heat flux sensor is an instrument which measures a heat flow passing through it by producing an electrical signal which is a linear function of the heat flow. These sensors are usually flat, thermally resistive plates (schematically shown in Figure~\ref{f:hfm}.) with thermocouples that have several thermophiles integrated inside of the thermocouple substrate. 

\subsection{Artificial neural networks}\label{s:ann}
\begin{figure}[h]
	\includegraphics[width=17pc]{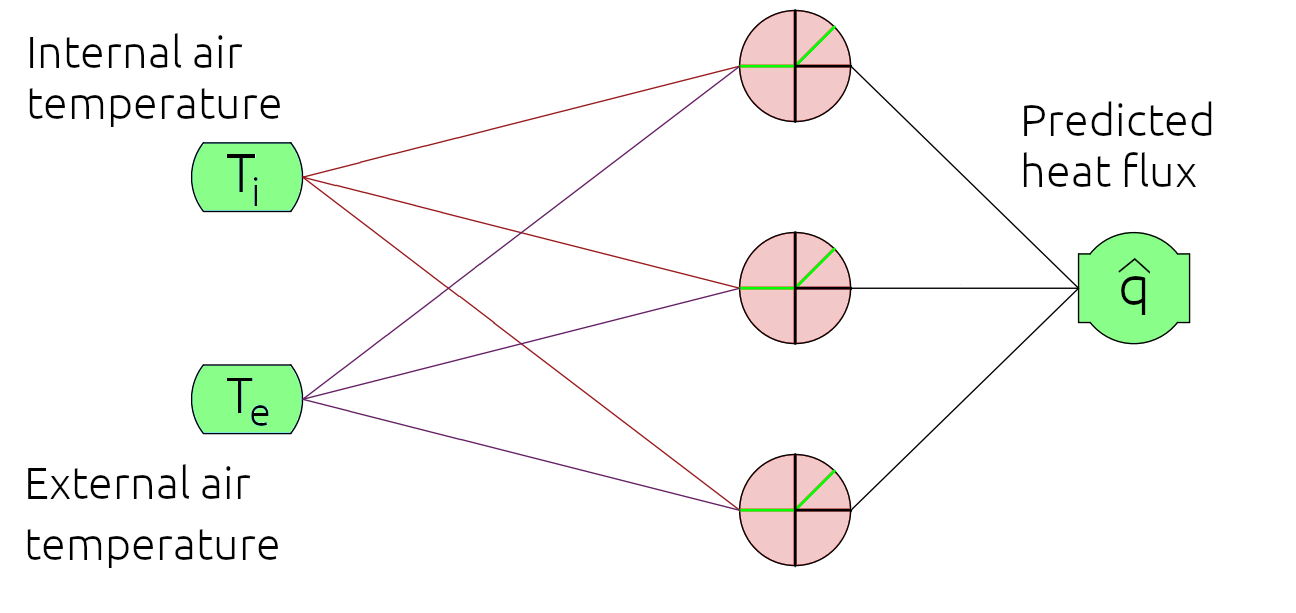}\hspace{2pc}%
	\begin{minipage}[b]{19pc}\caption{\label{f:mlp}Multilayer perceptron class of artificial neural network with two inputs, three neurons in hidden layer, and one output.}
	\end{minipage}
\end{figure}

Area of artificial intelligence and machine learning (ML) started to be used significantly in the last decade due to development of faster computer machines and engagement of new learning algorithms \cite{Jordan2015}. Large part of ML models depend on the concept of ANN. ANN is briefly a systematical set of weighting coefficients, biases and activation functions, and they behave as one function which transforms input variables to output ones with a goal to predict an outcome as close as possible to target values \cite{DL2021}. The most simple and firstly developed class of ANN is multilayer perceptron (MLP), schematically presented on figure \ref{f:mlp}. ANN model from figure \ref{f:mlp} is relatively simple and has two variables in the input layer, three neurons in hidden layer, and one output variable. Type of ANN model that has one output variable is called regression model compared to ANN model that has more outputs - classification model. The logic behind those two is the same, input variables are connected to neurons in hidden layers with functions that multiply each input and add a bias to it. In each neuron $i$ in hidden layer $n$, summation $z_i^{(n)}$ of transformed inputs is performed (\ref{eq:2}) and transformation of this summation with the activation function $\varphi_{n,i}$ (\ref{eq:3}) where $m$ is the number of inputs into neuron, $w_{0,n}^{(n)}$ is bias, $w_{j,i}^{(n)}$ is weighting coefficient and $y_i^{n}$ is value that is passed to the next layer, $n+1$.

\begin{center}
\begin{minipage}{14pc}
	\centering
	\begin{equation}\label{eq:2}
		z_{n,i} = w_{0,i}^{(n)} + \sum_{j=1}^{m} x_j w_{j,i}^{(n)}
	\end{equation}
\end{minipage} 
\begin{minipage}{14pc}
	\centering
	\begin{equation}\label{eq:3}
		y_{n,i} = \varphi_{n,i}(z_{n,i})
	\end{equation}
\end{minipage}
\end{center}

For the last layer $N$, i.e. output layer, and if we have only one variable $\hat{q}$, the predicted value is described with equation (\ref{eq:4}).

	\begin{equation}\label{eq:4}
		\hat{q} = w_0^{(N)} + \sum_{j=1}^{m_{N-1}} \varphi_j^{N-1}(z_{N-1,j}) \, w_{j}^{(N)}
	\end{equation}

Activation function is a nonlinear function which decides how much information is passed through each neuron to the next layer of neurons, and in this way information is carried from input layer through neurons in hidden layers all the way to output layer neurons. The most common activation functions are $sigmoid$, $tanh$ and $ReLU$.

ANN can learn by comparing predicted variables with actual ones. Firstly, it is essential to define loss $\mathcal{L}\left( f \left( x^{(i)}; \mathbf{W} \right), y^{(i)} \right)$ between predicted  $f \left( x^{(i)}; \textbf{W} \right)$ and actual $\left( y^{(i)} \right)$ variables where $\textbf{W}$ is matrix of weighting coefficients and biases. The empirical loss (\ref{eq:6}) can than be defined as total loss over the entire dataset. This is performed iteratively using the gradient method (often optimized) by computing the gradient $\frac{\partial J(\mathbf{W})}{\partial  \mathbf{W}}$ and by updating the weights matrix $\mathbf{W}$ until convergence is achieved.

\begin{equation}\label{eq:6}
	J(\mathbf{W}) = \frac{1}{n} \sum_{i=1}^{n} \mathcal{L}\left( f \left( x^{(i)}; \mathbf{W} \right), y^{(i)} \right)
\end{equation}

For regression problems, loss function $\mathcal{L}$ is usually mean squared error function. The "learning" is performed with minimization of empirical loss on network weights to obtain a matrix $\mathbf{W^*}$ with weighting coefficients that achieve the lowest loss (\ref{eq:7}).

\begin{equation}\label{eq:7}
	\mathbf{W^*} = \mathop{argmin}_{\mathbf{W}} \frac{1}{n} \sum_{i=1}^{n} \left( y^{(i)} - f \left( x^{(i)}; \mathbf{W} \right) \right)^2
\end{equation}

\begin{figure}[h]
	\centering
	\includegraphics[scale=0.145]{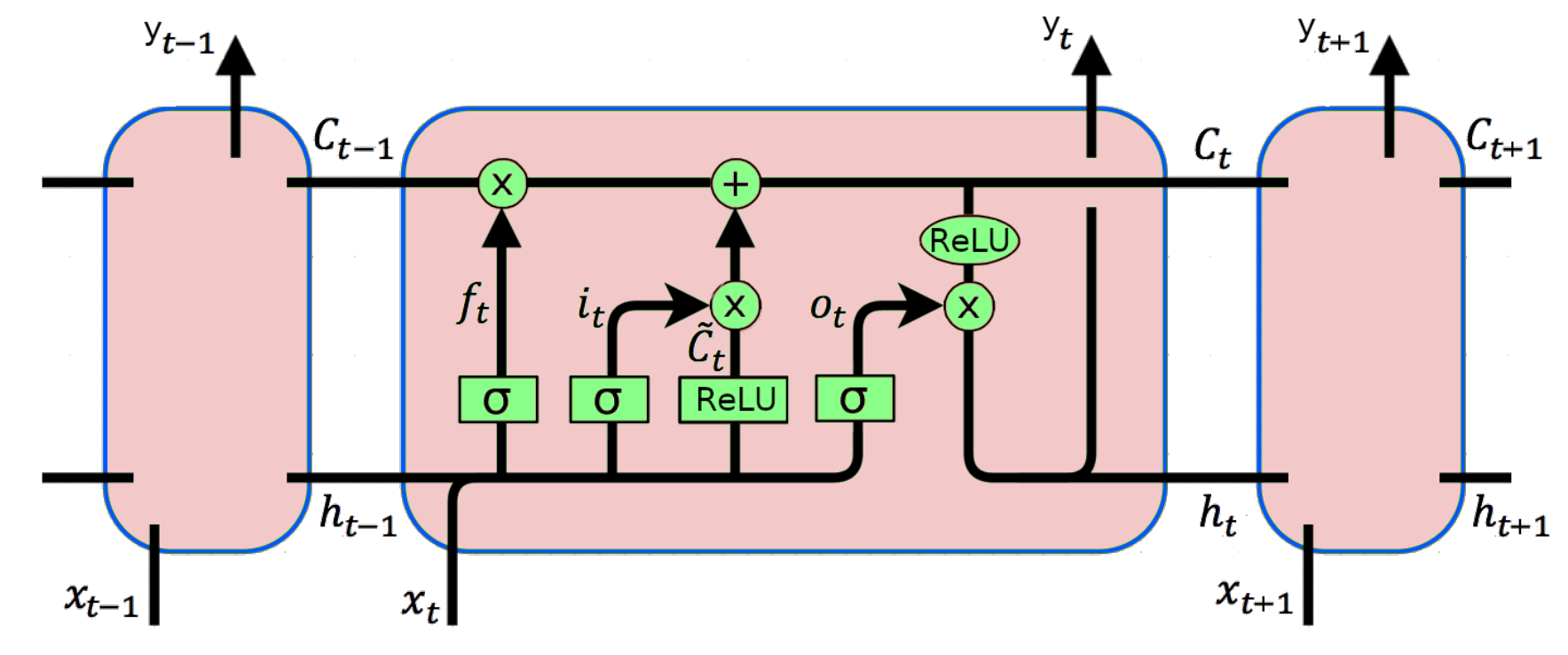}
	\caption{\label{f:lstm}Schematic view of long-short term memory unit with ReLU activation function.}
\end{figure}

For time series regression problems, another ANN class is commonly used -- recurrent neural networks (RNN). The reason for this is that RNN depend not only on input entry in certain timestep, but also on value from previous cell state. Two types of RNN were analyzed in the paper -- long-short term memory (LSTM) \cite{Hochreiter1997} cells and gated recurrent units (GRU) \cite{Cho2014}. Those two consist of set of gates that transform the inputs and previous cell states to provide outputs and transformed information that goes to the next cell. LSTM unit consists of four gates (figure \ref{f:lstm}): forget gate $f_t$ to forget irrelevant parts of the previous cell state, store gate $i_t$ for storing relevant new information into cell state, update for selective update of cell state values and output gate $o_t$ that controls what information is sent to the next timestep. 

\begin{figure}
	\centering
	\includegraphics[scale=0.155]{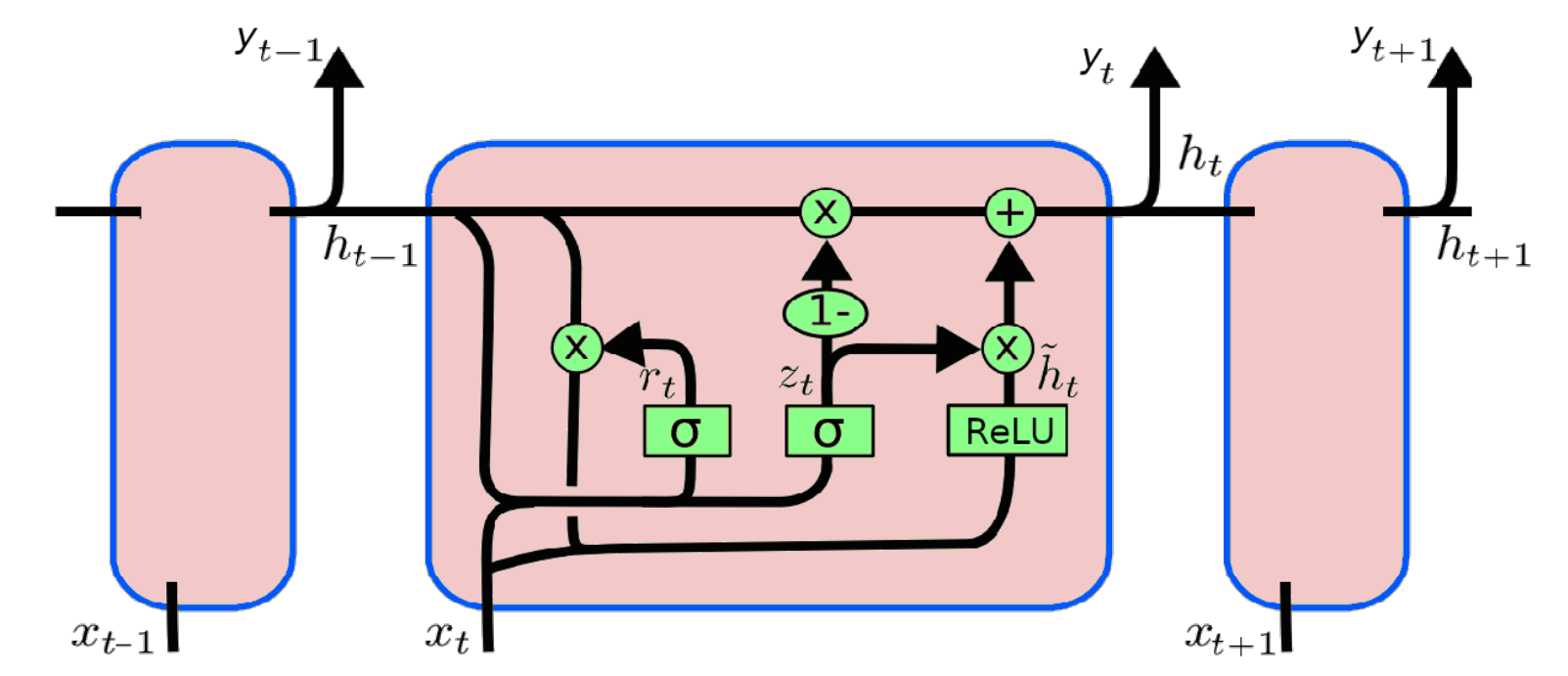}
	\caption{\label{f:gru}Schematic view of gated recurrent unit with ReLU activation function.}
\end{figure}

GRU has one gate less than LSTM (it does not have forget gate) and schematic view of it can be seen in figure \ref{f:gru}. It consists of reset gate $r_t$, update gate $z_t$ and candidate activation gate $\widetilde{h}_t$. Although they look slightly different, RNN follow the same principles described for MLP.

\subsection{Research design}

\begin{figure}[h]	
	\includegraphics[width=15pc]{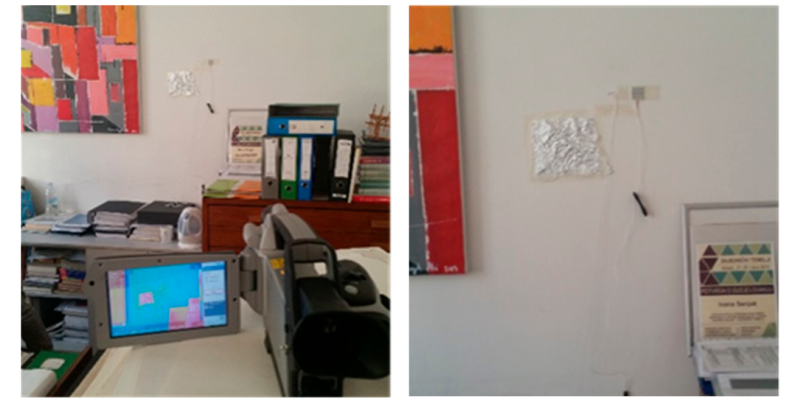} 
	\begin{minipage}[b]{21pc}\caption{\label{f:exp} Experimental setup for HFM data acquisition for analyzed wall.}
	\end{minipage}
\end{figure}

HFM experiment on a wall with unknown U-value was carried out in the period between 22 February 2019 and 26 February 2019 (figure \ref{f:exp}). Experimental data that is analyzed consists of 490 data entries collected every 10 minutes (in total 81.67 hours). In each time step, three values were measured -- internal and external air temperatures with thermocouples and heat flux with heat flux sensor. Prediction of heat flux is performed with four ANN architectures.

Four types of ANN architectures were analyzed: MLP3 -- multilayer perceptron with 3 neurons in hidden layer, LSTM100 -- architecture with 100 LSTM cells in hidden layer, GRU100 -- architecture with 100 GRU cells in hidden layer and LSTMGRU100 -- architecture with 50 LSTM cells in first hidden layer and 50 GRU cells in second hidden layer. Three cases of train/validation ratios were observed for each architecture: 1/4, 1/2 and 2/3. Results of predicted U-values were compared to the measured U-value. Predicted U-values were calculated in accordance with expression (\ref{eq:1}) applied on predicted results of heat flux. Point of interest was predicted sequence based on trained ML model. Comparison between predicted and true sequences was described by root mean squared error (RMSE), mean squared error (MSE) and mean absolute error (MAE).

\section{Results} \label{s:res}
Table \ref{t:results} shows information described in previous subsection with pointing up the best accuracy for certain train/validation ratio taking concern RMSE.

\begin{table}[h]
	\caption{\label{t:results}Results comparison of four types of ANN architectures. Prediction of heat flux (RMSE, MSE, MAE) and U-values -- prediction v.s. measured $0.586 \, W/(m^2 \, K)$.}
	\begin{center}
		\begin{tabular}{lcccccc}
			\br
			ANN type&train/validation&RMSE&MSE&MAE&predicted U-value&rel. difference\\
			\mr
			&1/4&2.02$^*$&4.06&1.44&0.534&8.96\%\\
			MLP3&1/2&2.14&4.58&1.55&0.574&2.23\%\\
			&2/3&1.91&3.63&1.41&0.546&7.03\%\\
			\mr
			&1/4&4.36&19.04&3.19&0.615&4.72\%\\
			LSTM100&1/2&1.68$^*$&2.82&1.14&0.556&5.31\%\\
			&2/3&1.52&2.31&1.03&0.546&6.99\%\\
			\mr
			&1/4&2.65&7.03&1.95&0.571&2.67\%\\
			GRU100&1/2&1.77&3.13&1.23&0.594&1.30\%\\
			&2/3&1.45&2.10&0.92&0.548&6.66\%\\
			\mr
			&1/4&2.88&8.31&1.98&0.573&2.30\%\\
			LSTMGRU100&1/2&2.02&4.08&1.22&0.605&3.02\%\\
			&2/3&1.22$^*$&1.48&0.68&0.563&4.13\%\\		
			\br
		\end{tabular}
		\raggedright \footnotesize $^*$ the best accuracy for certain train/validation ratio
	\end{center}
\end{table}

\begin{figure}[h]
	\includegraphics[width=19pc]{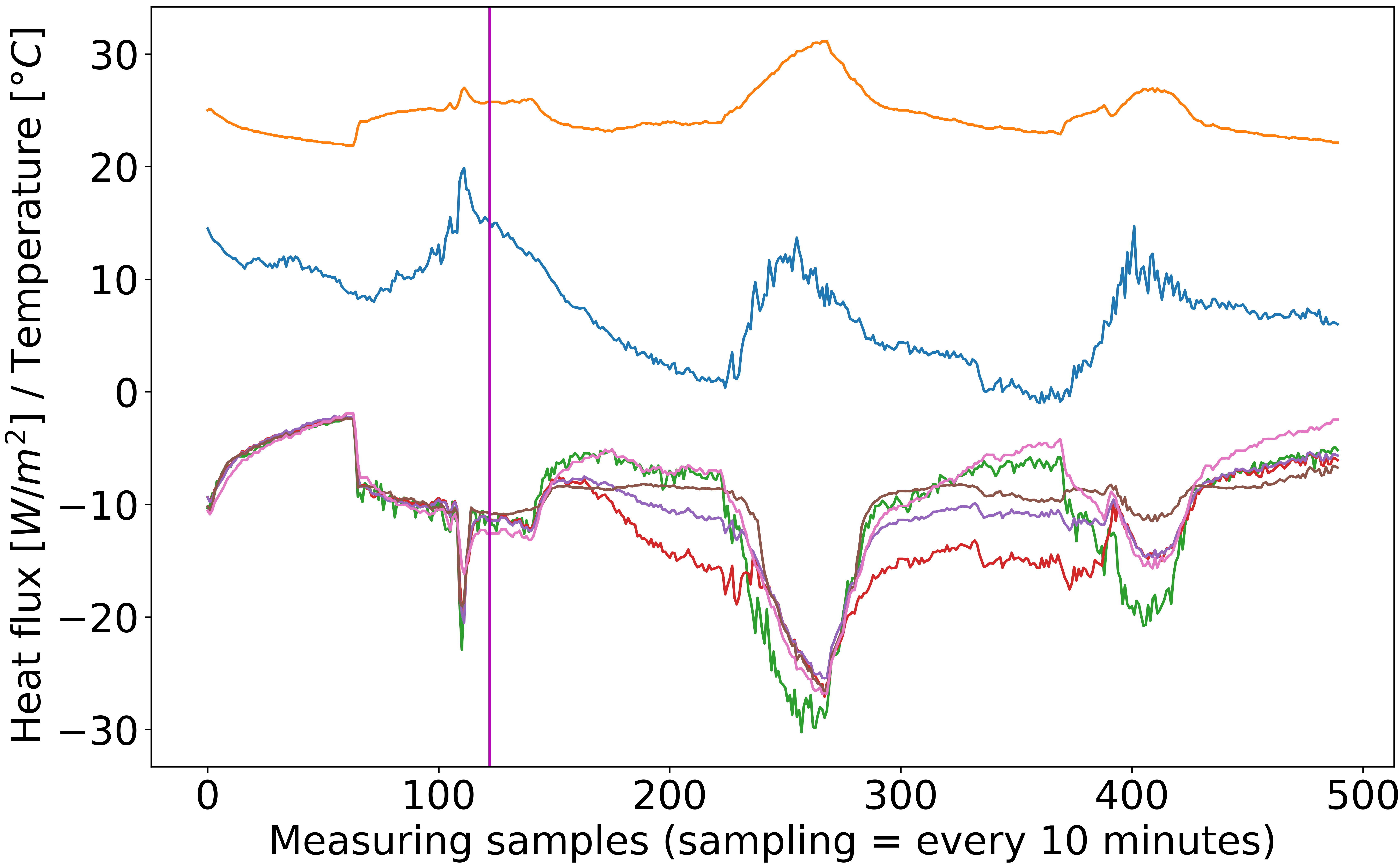}\hspace{1pc}%
	\begin{minipage}[b]{18pc}\includegraphics[width=10pc]{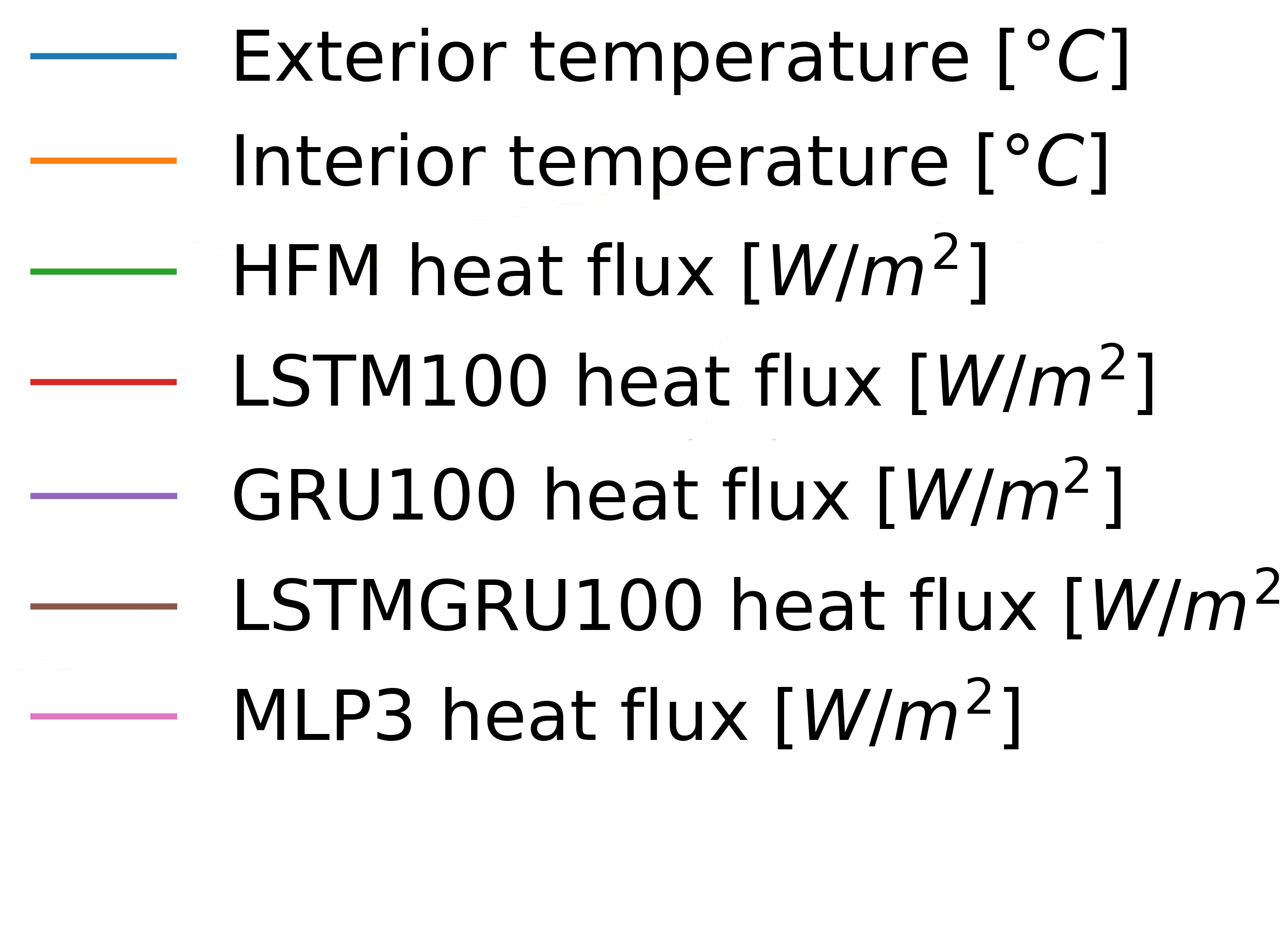} \\
		\caption{\label{f:e1_14}Results for training/validation ratio 1/4.}
	\end{minipage}
\end{figure}

\begin{figure}[h]
	\includegraphics[width=19pc]{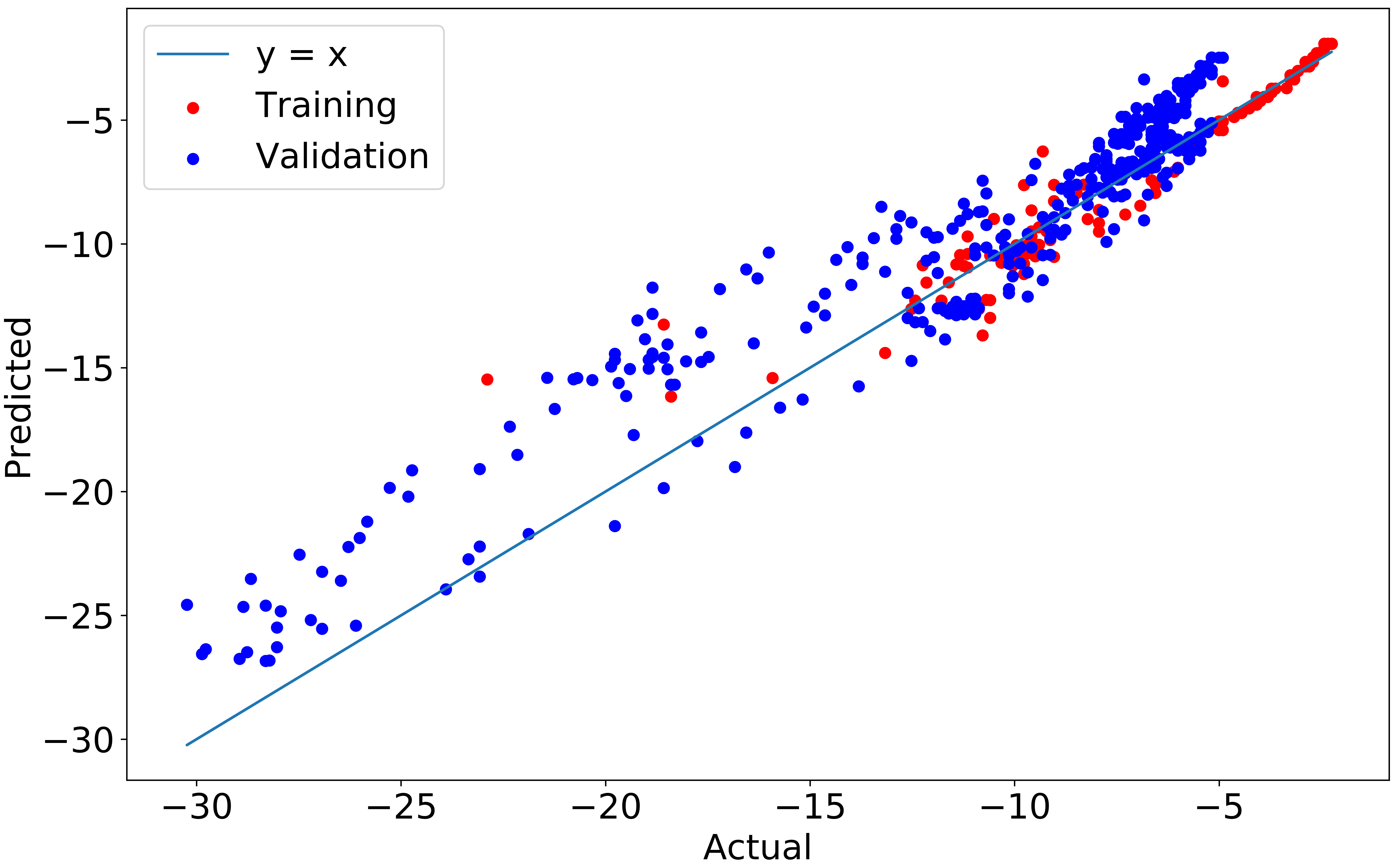}\hspace{1pc}%
	\begin{minipage}[b]{18pc}\caption{\label{f:e1_14_MLP3}Predicted vs. actual heat flux for MLP3 for training/validation ratio 1/4.}
	\end{minipage}
\end{figure}

\begin{figure}[h]
	\includegraphics[width=19pc]{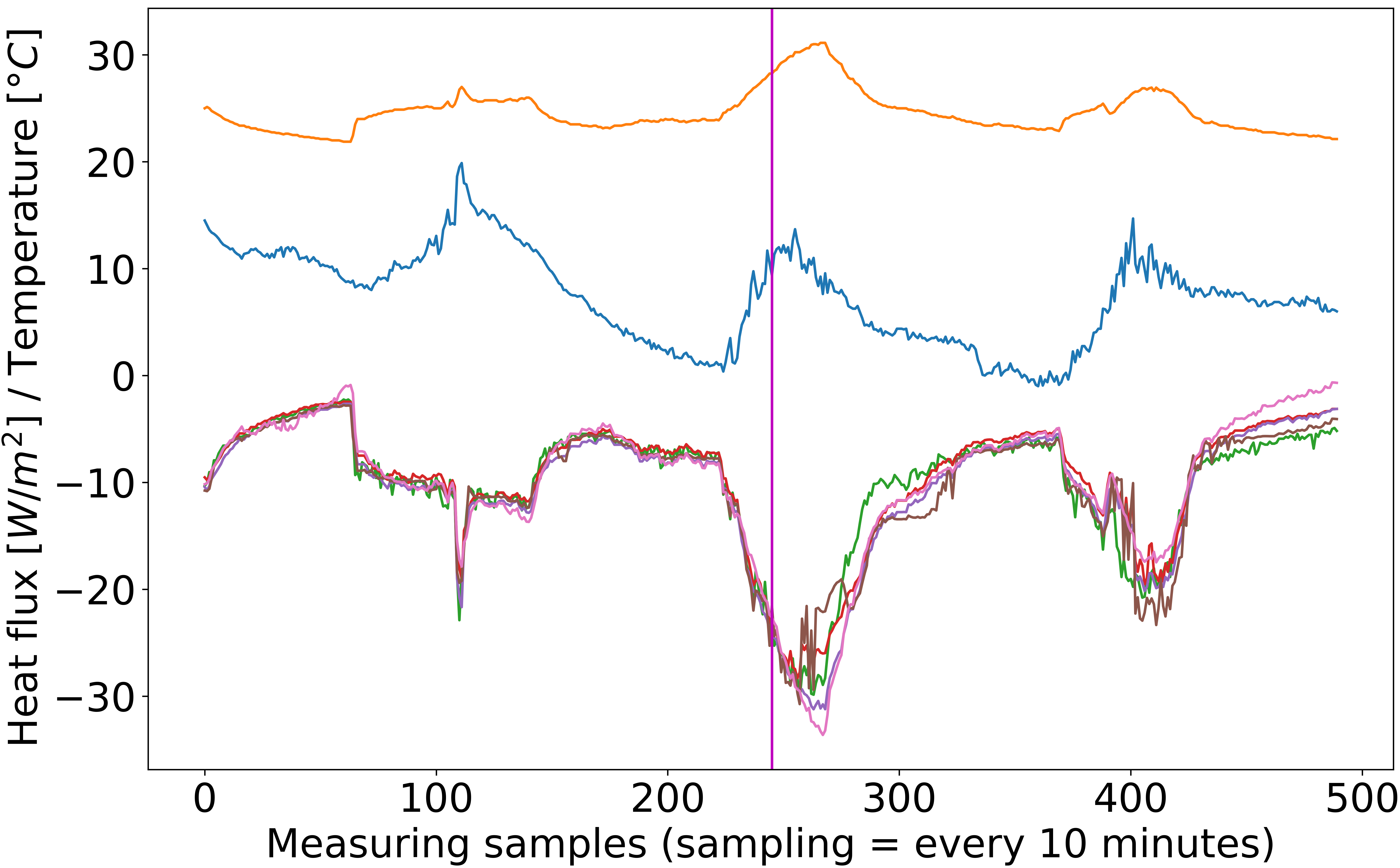}\hspace{1pc}%
	\begin{minipage}[b]{18pc}\includegraphics[width=10pc]{fig/legend.png} \\
		\caption{\label{f:e1_12}Results for training/validation ratio 1/2.}
	\end{minipage}
\end{figure}

\begin{figure}[h]
	\includegraphics[width=19pc]{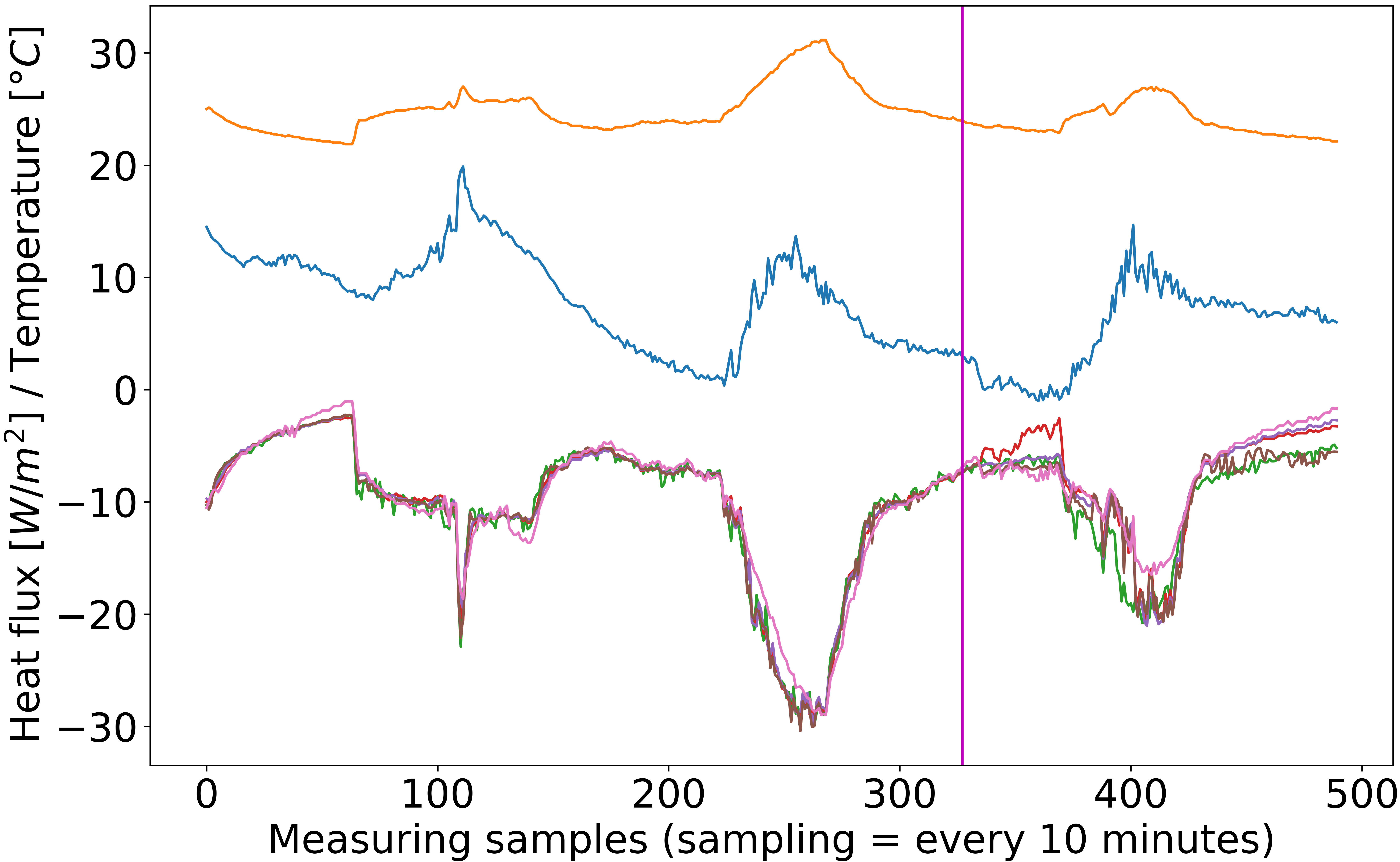}\hspace{1pc}%
	\begin{minipage}[b]{18pc}\includegraphics[width=10pc]{fig/legend.png} \\
		\caption{\label{f:e1_23}Results for training/validation ratio 2/3.}
	\end{minipage}
\end{figure}

If we observe training/validation ratio 1/4, the best matching is registered for MLP3 ANN architecture with RMSE $2.02$ which can be seen in table \ref{t:results}. Predictions from all analyzed ANN architectures for this case is shown in figure \ref{f:e1_14} and comparison of MLP3 predicted to actual (measured) values in figure \ref{f:e1_14_MLP3}. Vertical line is boundary between training and validation data. For training/validation ratios 1/2 and 2/3, better results were achieved as shown in table \ref{t:results} and on figures \ref{f:e1_12} and \ref{f:e1_23}. 

The best result is achieved for ANN architecture LSTMGRU100 for training/validation ratio 2/3 but it is probably because training part is 2/3 of whole dataset. More valuable result is for training/validation ratio 1/2 -- for ANN architecture LSTM100. RMSE for this ML model is $1.68$ and relative difference when comparing to measured U-value is $5.31 \%$.

\section{Discussion and conclusion}
The results are promising in terms of prediction of heat flux based on measured heat flux with two input temperatures in training period, and after the training period by measuring just input temperatures. Results show that RNN architectures give reliable results for larger dataset used for training (1/2 and 2/3 training/validation ratios) while MLP is better for smaller training dataset. This should be deeply analyzed because the goal is to reliably predict HFM results with small amount of data, but this prediction must be stable. MLP could not provide stability because it is not as nonlinear as RNN architectures so it might not reliably predict the HFM results in the long-run.  

\begin{figure}[h]
	\includegraphics[width=22pc]{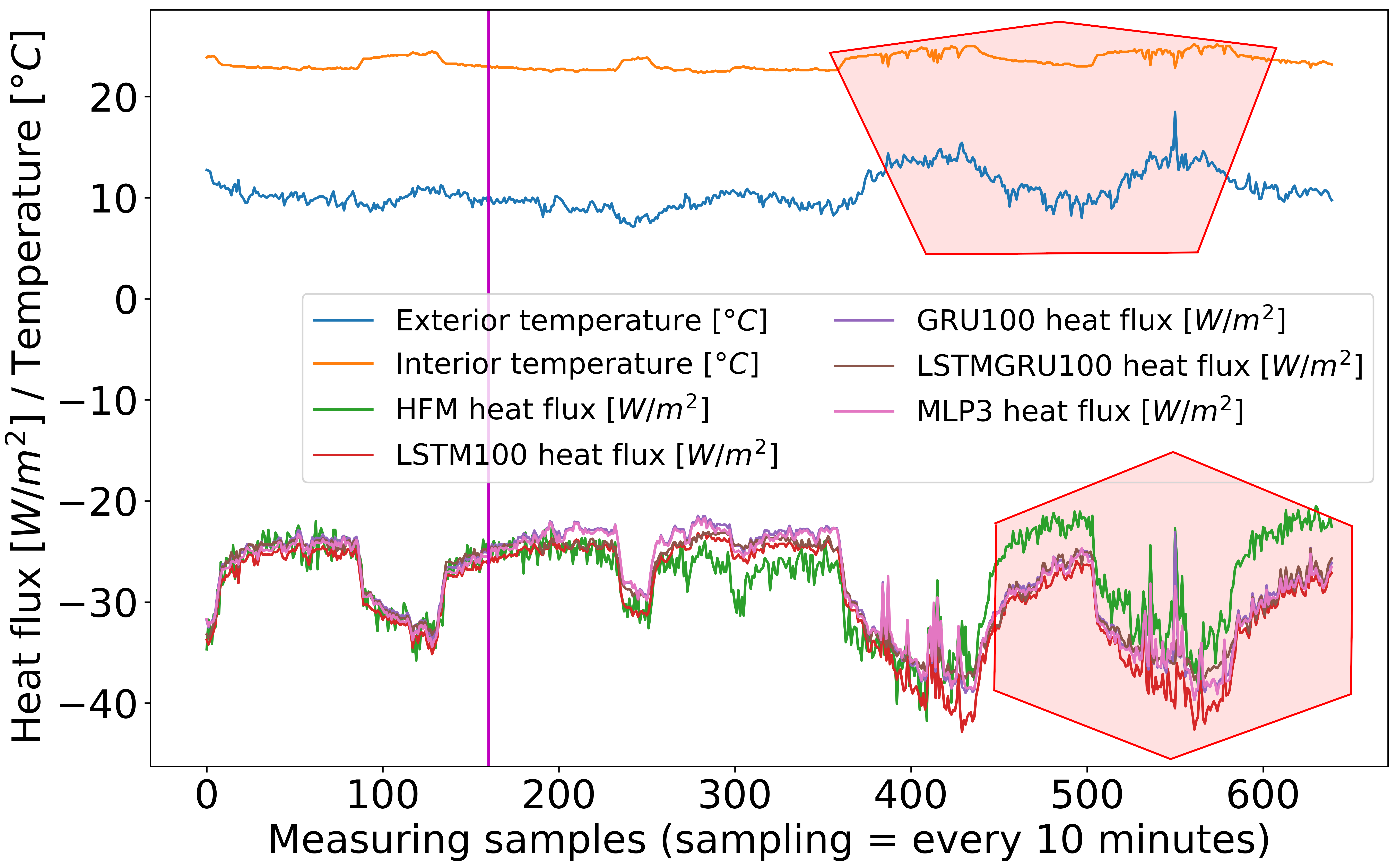}\hspace{1pc}%
	\begin{minipage}[b]{15pc}\caption{\label{f:e3_14}Analysis on additional wall with red-mark describing possible cause of non-stable prediction.}
	\end{minipage}
\end{figure}

The most noticeable limitation of the method could be an influence by a change in the boundary temperatures in a way it is not observed in ML training period. E.g. for wall in figure \ref{f:e3_14} it can be seen that after the training is done on 1/4 of total dataset, prediction is stable until such kind of change (red-marked in figure \ref{f:e3_14}). This can be solved by defining strict measuring conditions and/or by applying regularization techniques. For more certain conclusions, additional experimental investigation and analysis should be carried out.

\ack
One of the authors (Sanjin Gumbarević) would like to acknowledge the Croatian Science Foundation and European Social Fund for the support under the project ESF DOK-01-2018. This is preprint submitted to International Building Physics Conference 2021.

\section*{References}
\bibliography{BibTeX/iopart-num.bib}
\bibliographystyle{BibTeX/iopart-num.bst}

\end{document}